\documentclass[10pt,twocolumn,letterpaper]{article}
\pdfoutput=1 
\usepackage{cvpr}
\usepackage{times}
\usepackage{epsfig}
\usepackage{graphicx}
\usepackage{amsmath}
\usepackage{amssymb}
\usepackage[utf8x]{inputenc}

\usepackage{float}
\usepackage{algpseudocode}
\usepackage{siunitx}
\usepackage{lipsum}
\usepackage{algorithm}
\usepackage{caption}

\cvprfinalcopy

\begin{document}

\title{FIGR: Few-shot Image Generation with Reptile}

\author{Louis Clouâtre\\
École Polytechnique de Montréal\\
{\tt\small louis.clouatre@polymtl.ca}
\and
Marc Demers\\
McGill University\\
{\tt\small marc@cim.mcgill.ca}
}

\maketitle

\begin{abstract}

Generative Adversarial Networks (GAN) boast impressive capacity to generate realistic images. 
However, like much of the field of deep learning, they require an inordinate amount of data to produce results, thereby limiting their usefulness in generating novelty.
In the same vein, recent advances in meta-learning have opened the door to many few-shot learning applications. 
In the present work, we propose Few-shot Image Generation using Reptile (FIGR), a GAN meta-trained with Reptile.  
Our model successfully generates novel images on both MNIST and Omniglot with as little as 4 images from an unseen class.
We further contribute FIGR-8, a new dataset for few-shot image generation, which contains \num[group-separator={,}]{1548944} icons categorized in over \num[group-separator={,}]{18409} classes.
Trained on FIGR-8, initial results show that our model can generalize to more advanced concepts (such as ``bird'' and ``knife'') from as few as 8 samples from a previously unseen class of images and as little as 10 training steps through those 8 images.
This work demonstrates the potential of training a GAN for few-shot image generation and aims to set a new benchmark for future work in the domain.

\end{abstract}

\section{Introduction}
\vspace{0em}

Generative Adversarial Networks \cite{GAN} have helped bridge the gap between human and artificial intelligence with regard to understanding and manipulating images.
GANs however require several orders of magnitude more data points than humans in order to generate comprehensible images successfully from a given class of images.
This impairs the ability of GANs to generate novelty. In many cases, if the data is abundant enough to successfully train a GAN, there is little purpose to generating more of this data.

On the other hand, recent advances in meta-learning, like the MAML~\cite{MAML} and Reptile~\cite{Reptile} algorithms, have allowed learning tasks to perform well on novel data sampled from the same distribution as the training data.
These meta-learning algorithms have seen direct applications in supervised and reinforcement learning, but not in image generation.
Being very general in their application, those algorithms may be applicable to few-shot image generation. 
This paper defines the problem of few-shot image generation, and introduces an approach to GAN training for Few-shot Image Generation with Reptile (FIGR). 
In addition, this paper introduces FIGR-8, a dataset of \num[group-separator={,}]{1548944} black-and-white pictograms, ideograms, icons, emoticons, object or conception depictions categorized in \num[group-separator={,}]{18409} classes. We contribute this dataset as a challenging benchmark for one- and few-shot image generation approaches.
Following training, our approach is able to correctly generate images from a class of images with as few as $4$ samples from the previously unseen class. 
\vspace{0em}

In summary, our main contributions are:
\begin{itemize}
    \itemsep0em
    \item We develop a novel approach for training GANs for few-shot image generation.
    \item We contribute a challenging dataset for that same task.
\end{itemize}

The applications of few-shot image generation are broad, but we mainly foresee this approach to provide assistance in creative processes.
Artists or designers who lack time or creative inspiration for multiple versions of an image could sketch a limited number of drawings and have the trained model generate multiple similar versions of the sketches.

\section{Related work}
\subsection{Meta-learning}
MAML is currently the most widely used approach for few-shot meta-learning. Several variant of the algorithm exist. They all have conditions that make them ill-fitting for meta-training a GAN. First, they rely on the direction of the loss function to be linked with the quality of the model. For GAN's this assumption cannot be made. Second, they rely on being able to evaluate performance on a test set for training. There is no clear way to do that for GAN. 

\subsection{Few-Shot Image Generation}
\vspace{0em}

To our knowledge, Lake~\etal~(2015)~\cite{Omniglot} provides the first successful attempt at one-shot or few-shot image generation.
To achieve this on the Omniglot dataset introduced in the same paper, both the images and stroke data are used to train a Bayesian model through Bayesian~Program~Learning. It represents concepts, such as a pen stroke, as simple probabilistic programs and hierarchically combines them to generate images.
This yields a model that can be trained on a single image of a previously unseen letter and generate novel samples of the same letter.
It generates binary images.
\vspace{0em}

Rezende~\etal~(2016)~\cite{OneShotGen} uses a sequential generative model to achieve one-shot generation. The inference process uses an attention~\cite{Attention} module to have a Variational~Auto~Encoder~\cite{VAE} attend to a section of the generated image sequentially. Unlike in Lake~\etal~(2015), it trains on pure image data (without requiring stroke data), making this approach much more general. It generates binary images of size $28 \times 28$ and $52 \times 52$ on the Omniglot dataset with one-shot learning.
\vspace{0em}

Bartunov~and~Vetrov~(2018)~\cite{MatchingNetworksOneShot} uses matching networks to achieve few-shot image generation. In essence, matching networks~\cite{MatchingNetworks} are memory-assisted networks that leverage an external memory by employing an attention~\cite{Attention} module to quickly learn new concepts. It assumes that the concepts stored are somewhat similar to the new out-of-sample concepts. This approach is equally trained on pure image data and does not require a lengthy sequential inference period. It generates binary images of size $28 \times 28$ on the Omniglot dataset using few-shot learning.
\vspace{0em}

Several issues can be found with the aforementioned approaches that no prior work seems to address:
\begin{itemize}
    \item The use of small binary images for all generative models seem to imply scalability issues.  
    \item Limitations to the Omniglot dataset for one- and few-shot image generation. This dataset has several issues that will be expanded up in Section \ref{omni_related}
    \item None of the approaches have use an architecture that has shown the potential to generate highly realistic images like GANs have.
\end{itemize}

\subsection{Omniglot}
\label{omni_related}
The Omniglot~dataset~\cite{Omniglot} is the current baseline dataset for the one- or few-shot image generation task. Details about the dataset can be found in Section~\ref{omniglot_section}. There are two main issues with using this dataset as a benchmark.
\begin{itemize}
    \item All classes within the dataset are very similar. They all represent roughly the same concept-- a character. 
    \item The classes lack complexity. All classes in Omniglot are simple handwritten characters that can be explained and generated through the composition of learned pen strokes~\cite{Omniglot}. 
\end{itemize}
\vspace{0em}

We believe that a proper image generation benchmark should encompass a greater variety of classes and more complex classes to have real-life applications or the hope of applications on natural images.

\section{Few-shot Image Generation with Reptile}
\textbf{Generative Adversarial Networks} GANs are generative models that learn a generator network $G$ to map a random noise vector $z$ to an image $y$, such that $G(z) = y$. To accomplish this, we use a discriminator network $D$ and real images from the distribution we want to generate from $x$. $D$ is trained on both $x$ and $y$ to be able to distinguish the "fake" images $y$ from the "real" images $x$ while $G$ is trained to fool $D$. This adversarial game played between the two models leads to $G$ being able to generate images that resemble the ones from $x$~\cite{GAN}.
\vspace{0em}

\textbf{Few-shot image generation} We define the few-shot image generation problem with the help of the meta-learning problem set-up found in Finn~\etal~(2017)~\cite{MAML} and Nichol~\etal~(2018)~\cite{Reptile}. 
In this problem we assume access to a set of tasks $T$ containing multiple task $\tau$ where each individual task $\tau$ is an image generation problem with one class of images $X_\tau$ and a loss $L_\tau$. We define $L_\tau$ the ability of a human to discriminate between a group of generated images and a group of real images sampled from task $X_\tau$ as described in Lake~\etal~(2015)~\cite{Omniglot}.
We do not conduct human benchmarking in this paper as this will be part of follow up work. We however leave it in the task description as we believe it is essential for a proper metric to exist.

The aim is to find, through meta-training, parameters $\Phi$, that can quickly, meaning with little data and little training, converge on a random task $\tau$ to minimize an associated loss $L_\tau$.
\vspace{0em}

In essence, we want to:
\begin{equation} \label{reptile_}
minimize_{\Phi} \mathbb{E}_{\tau}[L_\tau(U_\tau^k(\Phi))]
\end{equation}

where $U_\tau^k(\Phi)$ is the operator that updates $\Phi$ $k$ times using $x_n$, a total of $n$ data points sampled from $X_\tau$~\cite{Reptile}.
\vspace{0em}

\textbf{MNIST} As an example, the MNIST dataset contains 10 classes (the 10 digits). In the few-shot image generation problem, they represent 10 tasks to solve, $\tau_0$ to $\tau_9$. We choose $\tau_0$ to $\tau_8$ to be the training task and $\tau_9$ to be the test task. Through meta-training on $\tau_0$ to $\tau_8$, we aim to obtain a set of parameters $\Phi$ that will quickly converge on a new $\tau$. We choose $n$ to be 4, meaning that we aim for our meta-trained $\Phi$ to converge to generating images of 9's with only 4 images sampled from $\tau_9$.
\vspace{0em}

\textbf{FIGR} In FIGR, $\Phi$ corresponds to both the generator network $G$ and the discriminator network $D$. $U$ corresponds to one step of Stochastic~Gradient~Descent~\cite{SGD}  on $D$ and $G$ using Wasserstein loss~\cite{WGAN} with gradient-penalty~\cite{WGAN-GP}. 
\vspace{0em}

The adapted Reptile pseudo code for meta-training the model is depicted in Algorithm~\ref{algo1}. The algorithm is composed of an outer loop and an inner loop. The inner loop is the $K$ step of the operator $U$ on a copy of the parameters $\Phi$ with task $\tau$. Once we have those adapted weight $W_\tau$, we can proceed to the outer loop. We set the gradient of $\Phi$ to be equal to $\Phi - W_\tau$. We then take one step with the Adam optimizer~\cite{Adam}.

\noindent
\begin{minipage}{\dimexpr\linewidth-1\fboxrule\relax}
\begin{algorithm}[H]\captionsetup{labelfont={sc,bf}, labelsep=newline, labelformat=empty}
  \caption{Algorithm 1: FIGR training}
  \label{algo1}
\begin{algorithmic}[1]
\State Initialize $\Phi_d$, the discriminator parameter vector
\State Initialize $\Phi_g$, the generator parameter vector
\For{iteration 1, 2, 3 ...} 
\State Make a copy of $\Phi_d$ resulting in $W_d$
\State Make a copy of $\Phi_g$ resulting in $W_g$
\State Sample task $\tau$
\State Sample $n$ images from $X_\tau$ resulting $x_\tau$ 
\For{$K > 1$ iterations}
\State Generate latent vector $z$
\State Generate fake images $y$ with $z$ and $W_g$
\State Perform step of SGD update on $W_d$ with\\\hspace{1.1cm} Wasserstein GP loss and $x_\tau$ and $y$
\State Generate latent vector $z$
\State Perform step of SGD update on $W_g$ with\\\hspace{1.1cm} Wasserstein loss and $z$ 
\EndFor
\State Set $\Phi_d$ gradient to be $\Phi_d$ - $W_d$
\State Perform step of Adam update on $\Phi_d$
\State Set $\Phi_g$ gradient to be $\Phi_g$ - $W_g$
\State Perform step of Adam update on $\Phi_g$
\EndFor
\end{algorithmic}
\end{algorithm}
\end{minipage}%

\vspace{1em}

Once meta-trained, we use a similar process to generate novel images from the sampled class described in Algorithm~\ref{algo2}.

\noindent
\begin{minipage}{\dimexpr\linewidth-2\fboxsep-2\fboxrule\relax}
\begin{algorithm}[H]\captionsetup{labelfont={sc,bf}, labelsep=newline, labelformat=empty}
\caption{Algorithm 2: FIGR generation}
\label{algo2}
\begin{algorithmic}[1]
\State Using $W_d$, a copy of the meta-trained $\Phi_d$
\State Using $W_g$, a copy of the meta-trained $\Phi_g$
\State Sample test task $\tau$
\State Sample $n$ images as $x_\tau$ from $X_\tau$ 
\For{$K >= 1$ iterations}
\State Generate latent vector $z$
\State Generate fake images $y$ with $z$ and $W_g$
\State Perform step of SGD update on $W_d$ with\\\hspace{0.6cm} Wasserstein GP loss and $x_\tau$ and $y$
\State Generate latent vector $z$
\State Perform step of SGD update on $W_g$ with\\\hspace{0.6cm} Wasserstein loss and $z$
\EndFor
\State Generate latent vector $z$
\State Generate fake images $y$
\end{algorithmic}
\end{algorithm}
\end{minipage}%

\vspace{1em}
For every task $\tau$ there exist optimal discriminator and generator weights $W_{d\tau}$ and $W_{g\tau}$.
Intuitively, Reptile initializes the weights $\Phi_d$ and $\Phi_g$ to the point in parameter space that minimizes the distance between $\Phi_d$, $\Phi_g$, $W_{d\tau}$ and $W_{g\tau}$ for all $\tau$, or 
\begin{equation}
minimize~{\sum\limits_{T} (\Phi_d - W_{d\tau}) + (\Phi_g - W_{g\tau})}
\label{reptile2}
\end{equation}

Hence, for a sampled task $\tau$, a model optimized with Reptile can quickly and with few data points converge to the optimal point $W_{d\tau}$, $W_{g\tau}$ from $\Phi_d$, $\Phi_g$. If the test tasks are close enough to the training task and if the training tasks are numerous enough, $\Phi_d$ and $\Phi_g$ are likely to be close to a test $\tau$'s $W_{d\tau}$ and $W_{g\tau}$.
This makes for rapid and easy generalization from few data points.
\vspace{0em}

\begin{figure*}[t]
\centering
\includegraphics[width=\textwidth]{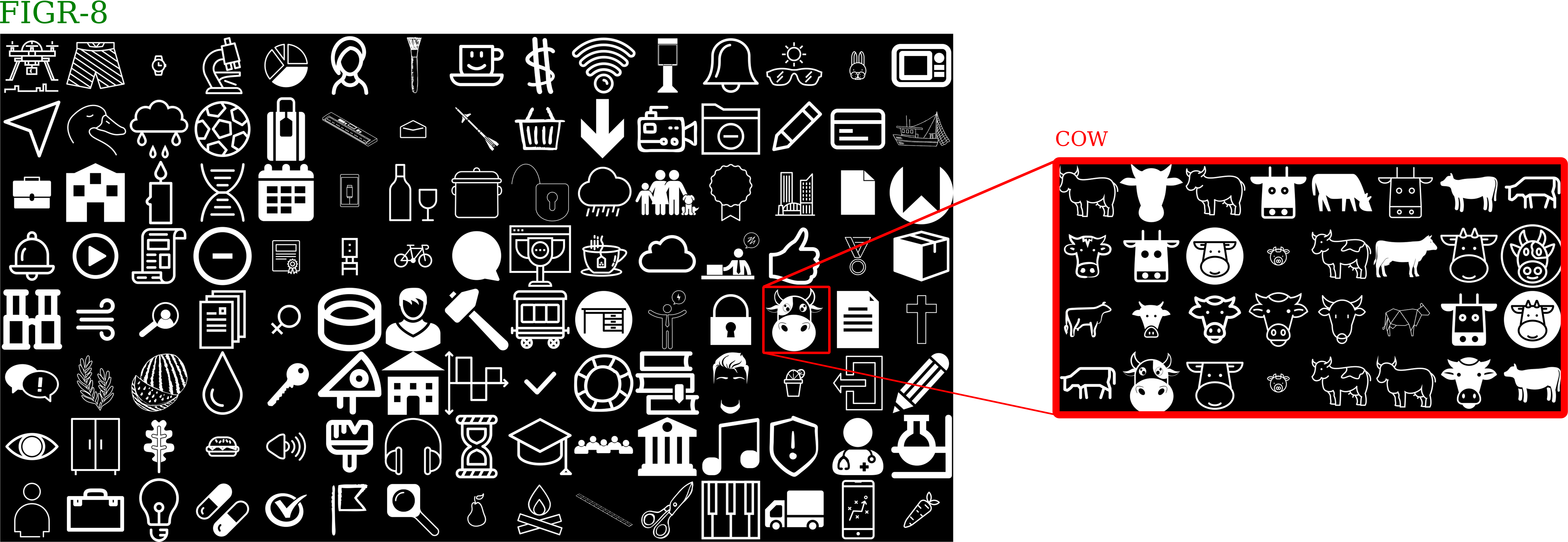}
\caption{Sample taken from the FIGR-8 dataset. Items from $120$ out of $18,409$ classes are displayed and one class (cow) is (non-extensively) detailed}
\label{fig:icon_dataset_example.png}
\end{figure*}

Reptile is broadly similar to joint training, and is effectively identical with a $K$ of 1.
However, by doing more gradient steps, we prioritize learning features that would be hard to reach, unlike joint training.
Assuming a 2D parameter space, a $K$ of 10 and a task $\tau$; a local minimum for parameter 1, $W_{\tau1}$, is reached after 2 gradient steps and a local minimum for the parameter 2, $W_{\tau2}$, is not reached after $K$ steps;
it is probable that:
\begin{equation}
\Phi_1 - W_{\tau1} < \Phi_2 - W_{\tau2}
\label{eq3}
\end{equation}
This would result in a larger outer loop update in the parameter space that is not readily attainable from $\Phi$ and smaller updates in the parameter space in which the model already possesses the ability to converge quickly.

\section{Datasets}

\subsection{MNIST} MNIST~\cite{Mnist} is the first dataset chosen as its simplicity allows us to iterate quickly through model ideas.
The MNIST dataset contains $28 \times 28$ grayscale images from the 10 digits.
We use the \num[group-separator={,}]{60000} training set images for all experiments.

\subsection{Omniglot}
\label{omniglot_section} 
Omniglot~\cite{Omniglot} is arguably the~{\it de~facto}~dataset for few-shot image generation. It contains $1623$ unique type of characters originating from 50 alphabets, each of which has been handwritten 1 time by 20 different individuals. 
Contrarily to MNIST, Omniglot allows for training our model on a much larger amount of classes of images, and test the out-of-sample performance of the model on a wider set of classes.

\subsection{FIGR-8} 
\begin{figure}[H]
\includegraphics[width=0.49\textwidth]{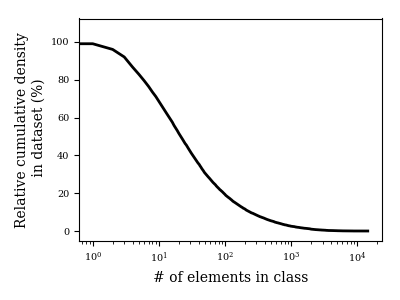}
\caption{Relative cumulative density of the number of elements in each class in the FIGR-8 dataset}
\label{fig:proportion_dataset.png}
\end{figure}

For the sake of testing the limits of our model, we compiled \num[group-separator={,}]{1548944} images separated in  \num[group-separator={,}]{18409} conceptually different classes, a set of data which we named FIGR-8. 
Each class contains at least $8$ images, up to a few thousands.
The icons are black-and-white representations of objects, concepts, patterns or designs that have been created by designers and artists and compiled into one data set. 
$120$ classes out of $18,409$ are pictured in Figure~\ref{fig:icon_dataset_example.png}. Each of those classes containing at least 8 images of a similar theme. Every image is of square format $1 \times 200 \times 200$.
The relative cumulative density of classes in the database is represented in Figure~\ref{fig:proportion_dataset.png}. 

We expect this dataset to be more challenging for training the meta-learning model, as it contains a wide variety of samples inside each class and a substantial amount of classes. Hopefully, the large amount of classes will let the model quickly understand the underlying concept even if every sample from a class does not represent the class' concept in the same manner. Some icons do have complex patterns and details, which poses a greater challenge than the existing datasets for one- or few-shot image generation tasks. All in all, the FIGR-8 dataset constitutes a tough yet achievable benchmark for few-shot image generation tasks.


\section{Experiments}
\subsection{Model architecture}
All models have been trained with Wasserstein loss~\cite{WGAN} with gradient-penalty~\cite{WGAN-GP}. We have found that a simple DCGAN~\cite{DCGAN} with a binary cross-entropy loss trained with this setup yielded positive results on MNIST~\cite{Mnist}. More complex datasets, such as Omniglot~\cite{Omniglot} and FIGR-8, were more challenging and required this loss function for the model to succeed. 
\vspace{0em}
Both the generator and the discriminator are built with residual neural networks~\cite{ResNet} with 18 layers. The discriminator uses layer normalization~\cite{LayerNorm} as prescribed in Gulrajani~\etal~(2017)~\cite{WGAN-GP}. The generator also uses layer normalization since batch normalization requires running statistics which are incompatible with Reptile's meta-update.

All rectified linear units are Parametric ReLU~\cite{PRELU} (PReLU). PReLU is the authors' preferred rectified linear activation function. However, any other rectified linear activation function should yield comparable results.

All images are resized with bilinear interpolation to $32 \times 32$ or $64 \times 64$.
All images are in grayscale format and normalized to have values constrained between $-1$ and $1$.
No data augmentation was used.
Results where sampled every $10,000$ meta-training steps and experiments took between $50,000$ and $250,000$ meta-training steps for results to converge.
All experiments were run on a single Tesla V100 on Google Cloud Platform (GCP). Training a model for $250,000$ meta-training steps with $n=4$ on Omniglot took $125$ hours with this setup. 
Table~\ref{tab:experiments} at the end of this paper shows hyperparameters for all experiments.

\subsection{Empirical Validation}
\vspace{0em}

In contrast with prior work, our model works on grayscale images rather than binary images. Our model also works without an external memory, a lengthy sequential inference process or additional training data in the form of pen stroke information. We believe that our approach, being built on top of GANs, has the best capacity to generalize to more challenging problems. 
\vspace{0em}

Shown below are the results of generating unseen test classes on our three datasets. The first row of every figure that follows represents the training data (circled in red). The following three rows are images generated by the model fine-tuned on those data points for 10 gradient steps. All images present results on previously unseen test classes. If unspecified, $n=4$.
\vspace{0em}

\textbf{MNIST} The MNIST data was rescaled to 32x32 pixel. The training classes are the digits from 0 to 8. The test class is the digit 9.  
\begin{figure}[!htb]
\includegraphics[width=0.45\textwidth]{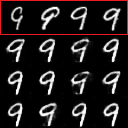}
\caption{MNIST; 50,000 update; 10 gradient steps}
\label{fig:MNIST_50k}
\end{figure}
On Figure~\ref{fig:MNIST_50k}, we can see good results on MNIST after 50,000 meta-training steps. This validates our approach on a toy problem.
\vspace{0em}

\textbf{Omniglot} The Omniglot data was resized to $32 \times 32$ and $64 \times 64$. The training classes where all $1623$ characters in the dataset minus $20$ randomly sampled character classes for the test set.

\begin{figure}[!htb]
\includegraphics[width=0.45\textwidth]{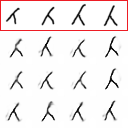}
\caption{Omniglot; 140,000 update; 10 gradient steps}
\label{fig:omniglot_32_good}
\end{figure}

\begin{figure}[!htb]
\includegraphics[width=0.45\textwidth]{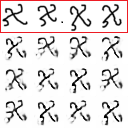}
\caption{Omniglot; 230,000 update; 10 gradient steps}
\label{fig:omniglot_32_bad}
\end{figure}

On simpler Omniglot characters like the one shown in Figure~\ref{fig:omniglot_32_good}, the model converges to good results after $140,000$ meta-training steps. On more complex characters, even after $230,000$ meta-training steps results are still lacking and humans can easily distinguish between most generated characters and the real ones. This is pictured in Figure~\ref{fig:omniglot_32_bad}.

\begin{figure}[h!]
\includegraphics[width=0.5\textwidth]{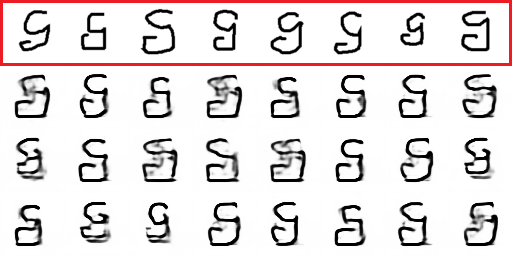}
\caption{Omniglot; 150,000 update; 10 gradient steps; $64 \times 64$; $n=8$}
\label{fig:omniglot_64}
\end{figure}

As for the $64 \times 64$ images, a batch size of $8$ was required to generate good results. In this case, after $150,000$ meta-training steps, around half the generated characters could conceivably fool a human judge. This is pictured in Figure~\ref{fig:omniglot_64}.
\vspace{0em}

\textbf{FIGR-8} The FIGR-8 data was resized to 32x32 pixels. The training classes where all $18,409$ classes minus $50$ randomly sampled classes for the test set. Here, $n=8$ was used for all experiments.

\begin{figure}[!h]
\includegraphics[width=0.5\textwidth]{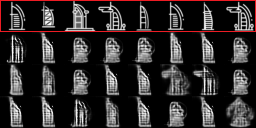}
\caption{FIGR-8; 80,000 update; 10 gradient steps; $n=8$}
\label{fig:figr-8_80k}
\end{figure}

\begin{figure}[!h]
\includegraphics[width=0.5\textwidth]{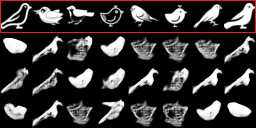}
\caption{FIGR-8; 90,000 update; 10 gradient steps; $n=8$}
\label{fig:figr-8_90k}
\end{figure}

\begin{figure}[b]
\includegraphics[width=0.5\textwidth]{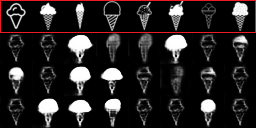}
\caption{FIGR-8; 100,000 update; 10 gradient steps; $n=8$}
\label{fig:figr-8_100k}
\end{figure}

For the FIGR-8 dataset, arguably none of the generated images pictured in Figures \ref{fig:figr-8_80k}, \ref{fig:figr-8_90k} and \ref{fig:figr-8_100k} can fool a human. We however see our model able to learn key features of the images very quickly, such as a birdlike shape or an ice cream cone.

\newpage
\section{Conclusion}
We have shown that Reptile can be used to effectively train Generative Adversarial Networks for few-shot image generation.
Using meta-training on a dataset containing several similar classes of images, we can learn to generate images from an unseen class with as little as $4$ samples on MNIST and Omniglot datasets.
This is done with no lengthy inference time, no external memory and no additional data. 
No hyperparameter tuning is required, the base parameters used are stable troughout experiments.
It is, to our knowledge, the first GAN trained for few-shot image generation.
Results show that our approach is able to quickly learn and generate simple concepts as well as complex ones.
Preliminary results on FIGR-8 show that a complex concept such as ``bird'' can be learned. 
To date, no other few-shot image generation model has managed to generate images other than handwritten characters.
The low amount of data required to generate images, once the model is pretrained, opens the door to several applications that were previously gated by the high amount of data required.
\vspace{0em}

We have also built, and will release for open source use, FIGR-8, a dataset containing over $18,409$ different classes and over $1,548,944$ images. 
Hopefully, this dataset will become a strong benchmark in the task of few-shot image generation.
\vspace{0em}

Several future directions should be explored:
\begin{itemize}
    \itemsep0em
    \item Generating multi-channel and/or larger images, such as with the CIFAR-100 dataset or the ImageNet dataset.
    \item Modifying batch normalization layers to be able to meta-train through them.
    \item Exploiting the wide variety of GAN architectures available.
    \item Using FIGR on ImageNet to make a pretrained GAN model for fine-tuning and transfer learning in the same capacity that ImageNet models are used for fine-tuning computer-vision models.
\end{itemize}

The code for the FIGR implementation can be found at https://github.com/OctThe16th/FIGR and the FIGR-8 database can be found at https://github.com/marcdemers/FIGR-8 and bit.ly/FIGR-8.


{\small
\bibliographystyle{plain}
\bibliography{egbib,md_bibliography}

\begin{thebibliography}{10}

\bibitem{WGAN}
Martin Arjovsky, Soumith Chintala, and L{\'e}on Bottou.
\newblock {W}asserstein generative adversarial networks.
\newblock In Doina Precup and Yee~Whye Teh, editors, {\em Proceedings of the
  34th International Conference on Machine Learning}, volume~70 of {\em
  Proceedings of Machine Learning Research}, pages 214--223, International
  Convention Centre, Sydney, Australia, 06--11 Aug 2017. PMLR.

\bibitem{LayerNorm}
Lei~Jimmy Ba, Ryan Kiros, and Geoffrey~E. Hinton.
\newblock Layer normalization.
\newblock {\em CoRR}, abs/1607.06450, 2016.

\bibitem{Attention}
Dzmitry Bahdanau, Kyunghyun Cho, and Yoshua Bengio.
\newblock Neural machine translation by jointly learning to align and
  translate.
\newblock {\em arXiv e-prints}, abs/1409.0473, September 2014.

\bibitem{MatchingNetworksOneShot}
Sergey Bartunov and Dmitry~P. Vetrov.
\newblock Few-shot generative modelling with generative matching networks.
\newblock In {\em AISTATS}, 2018.

\bibitem{SGD}
L{\'e}on Bottou.
\newblock Large-scale machine learning with stochastic gradient descent.
\newblock In Yves Lechevallier and Gilbert Saporta, editors, {\em Proceedings
  of COMPSTAT'2010}, pages 177--186, Heidelberg, 2010. Physica-Verlag HD.

\bibitem{MAML}
Chelsea Finn, Pieter Abbeel, and Sergey Levine.
\newblock Model-agnostic meta-learning for fast adaptation of deep networks.
\newblock {\em CoRR}, abs/1703.03400, 2017.

\bibitem{GAN}
Ian Goodfellow, Jean Pouget-Abadie, Mehdi Mirza, Bing Xu, David Warde-Farley,
  Sherjil Ozair, Aaron Courville, and Yoshua Bengio.
\newblock Generative adversarial nets.
\newblock In Z.~Ghahramani, M.~Welling, C.~Cortes, N.~D. Lawrence, and K.~Q.
  Weinberger, editors, {\em Advances in Neural Information Processing Systems
  27}, pages 2672--2680. Curran Associates, Inc., 2014.

\bibitem{WGAN-GP}
Ishaan Gulrajani, Faruk Ahmed, Martin Arjovsky, Vincent Dumoulin, and Aaron
  Courville.
\newblock Improved training of wasserstein gans, 2017.

\bibitem{PRELU}
Kaiming He, Xiangyu Zhang, Shaoqing Ren, and Jian Sun.
\newblock Delving deep into rectifiers: Surpassing human-level performance on
  imagenet classification.
\newblock {\em 2015 IEEE International Conference on Computer Vision (ICCV)},
  Dec 2015.

\bibitem{ResNet}
Kaiming He, Xiangyu Zhang, Shaoqing Ren, and Jian Sun.
\newblock Deep residual learning for image recognition.
\newblock {\em 2016 IEEE Conference on Computer Vision and Pattern Recognition
  (CVPR)}, Jun 2016.

\bibitem{Adam}
Diederik~P. Kingma and Jimmy Ba.
\newblock Adam: A method for stochastic optimization, 2014.

\bibitem{VAE}
Diederik~P. Kingma and Max Welling.
\newblock Auto-encoding variational bayes.
\newblock {\em CoRR}, abs/1312.6114, 2013.

\bibitem{Omniglot}
Brenden~M. Lake, Ruslan Salakhutdinov, Jason Gross, and Joshua~B. Tenenbaum.
\newblock One shot learning of simple visual concepts.

\bibitem{Mnist}
Yann LeCun and Corinna Cortes.
\newblock {MNIST} handwritten digit database.
\newblock 2010.

\bibitem{Reptile}
Alex Nichol, Joshua Achiam, and John Schulman.
\newblock On first-order meta-learning algorithms, 2018.

\bibitem{DCGAN}
Alec Radford, Luke Metz, and Soumith Chintala.
\newblock Unsupervised representation learning with deep convolutional
  generative adversarial networks, 2015.

\bibitem{OneShotGen}
Danilo Rezende, Shakir, Ivo Danihelka, Karol Gregor, and Daan Wierstra.
\newblock One-shot generalization in deep generative models.
\newblock In Maria~Florina Balcan and Kilian~Q. Weinberger, editors, {\em
  Proceedings of The 33rd International Conference on Machine Learning},
  volume~48 of {\em Proceedings of Machine Learning Research}, pages
  1521--1529, New York, New York, USA, 20--22 Jun 2016. PMLR.

\bibitem{MatchingNetworks}
Oriol Vinyals, Charles Blundell, Tim Lillicrap, koray kavukcuoglu, and Daan
  Wierstra.
\newblock Matching networks for one shot learning.
\newblock In D.~D. Lee, M.~Sugiyama, U.~V. Luxburg, I.~Guyon, and R.~Garnett,
  editors, {\em Advances in Neural Information Processing Systems 29}, pages
  3630--3638. Curran Associates, Inc., 2016.

\end{thebibliography}
}

\begin{table*}[t]
\centering
\begin{tabular}{ |c|c|c|c| } 
 \hline
   & \textbf{MNIST} & \textbf{Omniglot} & \textbf{FIGR-8} \\ 
 \hline
 \textbf{Inner learning rate} & 0.0001 & 0.0001 & 0.0001 \\ 
 \hline
 \textbf{Outer learning rate} & 0.00001 & 0.00001 & 0.00001 \\ 
 \hline
 \textbf{Training size n} & 4 & 4 and 8 & 8 \\ 
 \hline
  \textbf{Inner loops K} & 10 & 10 & 10 \\ 
 \hline
 \textbf{Image resize} & $32 \times 32$ & $32 \times 32$ and $64 \times 64$ & $32 \times 32$ \\
 \hline
 \textbf{Grayscale} & True & True & True \\
 \hline
 \textbf{Validation classes} & 1 & 20 & 50 \\
 \hline
\end{tabular}
\caption{Hyperparameters for all experiments}
\label{tab:experiments}
\end{table*}

\end{document}